%% file: sample-sigconf-authordraft.tex
\documentclass[sigconf,balance=false]{acmart}
\usepackage{amsmath}
\usepackage{algorithmic}
\usepackage{algorithm}
\usepackage{array}
\usepackage[caption=false,font=normalsize,labelfont=sf,textfont=sf]{subfig}
\usepackage{textcomp}
\usepackage{url}
\usepackage{verbatim}
\usepackage{graphicx}
\usepackage{lineno}
\usepackage{multirow}
\usepackage{tabularx}
\usepackage{cancel}
\usepackage{caption}
\usepackage{pgfplots}
\pgfplotsset{compat=newest}
\usepackage{tikz}
\usepackage{float}
\usepackage{balance}
\usepackage{etoolbox}
\usepackage{booktabs}
\usepackage{microtype}
\usepackage{chngcntr}
\usepackage{utfsym}
\usepackage[table]{xcolor}

\AtBeginDocument{%
  \providecommand\BibTeX{{%
    \normalfont B\kern-0.5em{\scshape i\kern-0.25em b}\kern-0.8em\TeX}}}

\copyrightyear{2026}
\acmYear{2026}
\setcopyright{cc}
\setcctype{by}
\acmConference[MM '26]{Proceedings of the 34th ACM International Conference on Multimedia}{November 10--14, 2026}{Rio de Janeiro, Brazil}
\acmBooktitle{Proceedings of the 34th ACM International Conference on Multimedia (MM '26), November 10--14, 2026, Rio de Janeiro, Brazil}
\acmDOI{10.1145/3767308.3836045}
\acmISBN{979-8-4007-2213-4/2026/11}

\title{Beyond Token Stacking: Multi-scale Visual Modulation and Adaptive Fused-feature Calibration for MLLMs}

\author{Ye Xiao}
\affiliation{%
  \institution{National University of Defense Technology}
  \city{Changsha}
  \country{China}
}
\email{xiaoye1120@nudt.edu.cn}

\author{Shezheng Song}
\authornote{Corresponding authors.}
\affiliation{%
  \institution{Hefei University of Technology}
  \city{Hefei}
  \country{China}
}
\email{betterszsong@gmail.com}

\author{Yi Ren}
\affiliation{%
  \institution{National University of Defense Technology}
  \city{Changsha}
  \country{China}
}
\email{renyi@nudt.edu.cn}

\author{Shasha Li}
\affiliation{%
  \institution{National University of Defense Technology}
  \city{Changsha}
  \country{China}
}
\email{shashali@nudt.edu.cn}

\author{Bin Ji}
\affiliation{%
  \institution{National University of Defense Technology}
  \city{Changsha}
  \country{China}
}
\email{jibin@nudt.edu.cn}

\author{Xiaodong Liu}
\affiliation{%
  \institution{National University of Defense Technology}
  \city{Changsha}
  \country{China}
}
\email{liuxiaodong@nudt.edu.cn}

\author{Jun Ma}
\affiliation{%
  \institution{National University of Defense Technology}
  \city{Changsha}
  \country{China}
}
\email{majun@nudt.edu.cn}

\author{Jie Yu}
\authornotemark[1]
\affiliation{%
  \institution{National University of Defense Technology}
  \city{Changsha}
  \country{China}
}
\email{yj@nudt.edu.cn}


\begin{abstract}
  Current Multimodal Large Language Models (MLLMs) have advanced rapidly and follow two main paradigms: one involves concatenating visual and textual tokens for input into Large Language Model (LLM), while the other injects visual knowledge into Feed-Forward Networks (FFNs) of the LLM. However, the former introduces a large number of redundant visual tokens, substantially increasing computational overhead. The latter reduces sequence length but removes images from the input stage entirely, which may hinder the activation of image-related knowledge. Moreover, naively injecting visual knowledge into the FFNs could induce activation distribution shifts due to cross-modal semantic discrepancies. To address these challenges, we develop \textbf{B}eyond \textbf{T}oken \textbf{S}tacking (BTS), a novel visual enhancement framework for MLLMs. Specifically, we propose a Multi-scale Visual Modulation (MVM) mechanism that injects multi-scale visual information at the input stage via modulation without increasing the token sequence length. Furthermore, we develop an Adaptive Fused-feature Calibration (AFC) module that performs per-channel affine calibration to decouple visual injection from distribution alignment in the FFN layers, improving multimodal fusion stability while preserving language priors. Experiments show that our approach achieves state-of-the-art performance. \footnote{Code: ~\href{https://github.com/November198/BTS}{https://github.com/November198/BTS}}
  
\end{abstract}


\begin{document}
\maketitle

\begin{figure*}[h!] 
\centering
\includegraphics[width=\linewidth,trim=130 300 100 140,clip]{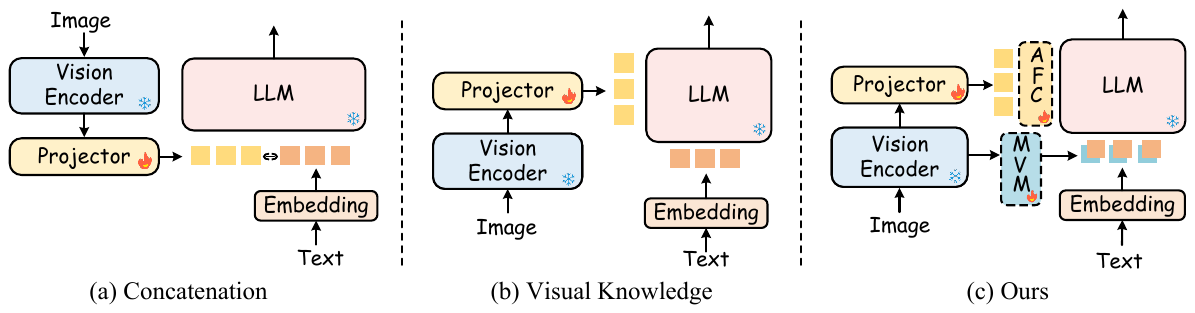} 
\color{blue}
\vspace{-0.65cm} 
\caption{Framework comparisons between previous methods and our method.}
\label{fig:1}
\vspace{-0.2cm} 
\end{figure*}

\section{Introduction}
In recent years, the rapid advancement of LLMs \cite{touvron2023llama,achiam2023gpt} has significantly propelled the evolution of MLLMs \cite{hurst2024gpt,dai2023instructblip}, enabling them to handle multiple modalities (\textit{i.e.}, text, images, video, and audio). Notably, Vision-Language Models (VLMs) \cite{song2025bridge,song2024dual,wang2024qwen2}, which integrate pretrained visual encoders with LLMs, have achieved impressive results on a range of multimodal tasks, including image caption \cite{ke2019reflective} and visual question answering \cite{antol2015vqa}. 
 
To effectively align and fuse visual and linguistic representations, VLMs typically rely on full-parameter fine-tuning. However, as the parameter scales of both visual and language models grow exponentially, full fine-tuning that updates most or all parameters requires substantial training memory and computational costs. To mitigate this challenge, researchers have explored parameter-efficient fine-tuning (PEFT) for VLMs \cite{luo2023cheap,jie2024memory,sung2022vl,hu2023vl,chen2022adaptformer,hao2025adem}, where pretrained backbones are largely kept frozen and only a small set of lightweight modules is updated. Depending on how visual information is incorporated into the LLM, existing methods can be categorized into two paradigms, as illustrated in Figure~\ref{fig:1}.

\textit{(i) Concatenation}: This paradigm \cite{luo2023cheap,sung2022vl,hu2023vl} projects the output of the visual encoder into the input space of the language model via linear projection, then concatenates the visual prompts with text embedding tokens, and finally feeds the combined sequence as a whole into the LLM. Although the above methods preserve relatively rich visual information, the large number of visual tokens often exhibits substantial redundancy. For example, an image encoded by ViT/14 produces up to 196 patch tokens, whereas the accompanying text is typically far shorter, averaging 12.84 tokens on GQA \cite{hudson2019gqa} and 6.3 tokens on VQAv2 \cite{goyal2017making}.
\textit{(ii) Visual Knowledge}: This paradigm \cite{jie2024memory,gao2023llama,hao2025adem} injects visual knowledge into the FFN layers of the LLM to avoid input sequence-length inflation. However, these approaches \textbf{lack visual contextual information at the input stage}, which may reduce the likelihood of activating image-related knowledge. 
Meanwhile, the injected visual features are usually extracted at a single granularity, making it difficult to capture both global structure and fine-grained details. In fact, robust visual grounding often benefits from multi-scale information that captures both global semantics and fine-grained details. 
Additionally, due to the inherent \textbf{semantic gap and mismatched feature statistics} between vision and language \cite{yin2024survey}, naively injecting visual features into the FFN layers could perturb activation distributions, inducing distribution shifts and compromising the stability of cross-modal feature fusion \cite{ioffe2015batch}.

To address the aforementioned issues, in this paper, we develop a novel visual enhancement framework for MLLMs named BTS. Our method consists of two key components: \textbf{(1) Multi-scale Visual Modulation (MVM)}: Since FFN-based visual knowledge injection methods lack explicit and rich visual contextual priors at the input stage, we propose MVM mechanism. Specifically, we perform multi-stage downsampling on the visual encoder outputs to extract multi-scale visual information that integrates coarse-grained semantics with fine-grained details. We then employ a lightweight adapter to modulate these multi-scale features into a visual bias, which is injected into text token embeddings in a residual manner. This design introduces multi-scale visual information into the language model without increasing the token sequence length. \textbf{(2) Adaptive Fused-feature Calibration (AFC)}: To mitigate the activation distribution perturbations induced by visual knowledge injection in the FFN layers, we introduce AFC module. AFC adaptively calibrates fused features via per-channel affine transformations to decouple visual injection from distribution alignment, where the injection branch contributes cross-modal gains and the calibration branch restores activation statistics to a stable and controllable regime. Consequently, AFC achieves a more reliable multimodal fusion process and maintains the linguistic priors of the LLM.
We conduct systematic evaluations of BTS on multiple multimodal benchmarks, and extensive experimental results show that BTS significantly enhances the multimodal understanding capability of the model. Overall, our contributions are summarized as follows:

\begin{itemize}
  \item We propose a Multi-scale Visual Modulation (MVM) mechanism, which leverages lightweight feature modulation to achieve efficient injection of multi-scale visual information at the input stage, thereby enhancing the ability of the model to perceive both scene structure and fine-grained details.
  \item We introduce an Adaptive Fused-feature Calibration (AFC) module that performs adaptive calibration of cross-modal fused features via per-channel affine transformations. This module effectively mitigates activation distribution shifts induced by visual knowledge injection and improves the effectiveness of multimodal feature fusion.
  \item We conduct extensive experiments on multiple multimodal benchmarks. BTS achieves state-of-the-art performance, demonstrating its effectiveness.
\end{itemize}

\begin{figure*}[h!] 
\centering
\includegraphics[]{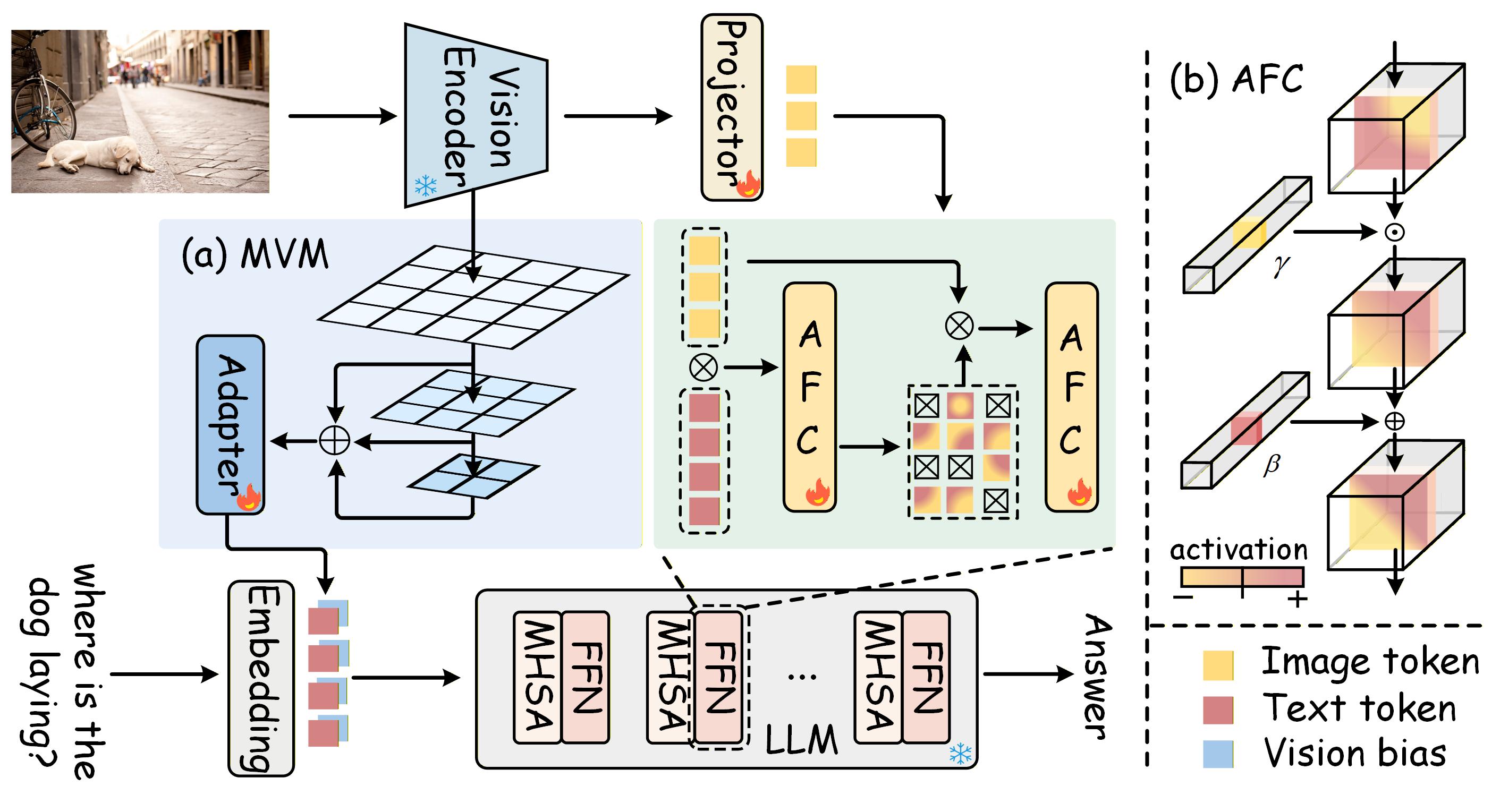}
\vspace{-0.4cm}
\caption{Method overview. BTS consists of two principal components: a Multi-scale Visual Modulation (MVM) mechanism and an Adaptive Fused-feature Calibration (AFC) module. MVM injects multi-scale visual contextual information into text embeddings via lightweight residual modulation without increasing the token sequence length, whereas AFC calibrates cross-modal fused features by applying per-channel affine transformations in the FFN layers, thereby stabilizing cross-modal fusion.}
\label{fig:2}
\vspace{-0.2cm}
\end{figure*}


\section{Related Work}
\subsection{Multimodal Large Language Models}
Multimodal Large Language Models unify modalities such as vision and language within a single framework, enabling end-to-end perception and reasoning. Early work primarily relied on cross-modal contrastive learning (\textit{e.g.}, CLIP \cite{radford2021learning} and ALBEF \cite{li2021align}) or encoder-decoder pre-training architectures (\textit{e.g.} ViLBERT \cite{lu2019vilbert}, LXMERT \cite{tan2019lxmert}, and Unified VLP \cite{zhou2020unified}) to learn vision-language representations. Recently, an increasing number of studies have shifted toward efficiently integrating visual information into LLMs to support image understanding, multi-turn dialogue, and complex reasoning. For instance, Flamingo \cite{alayrac2022flamingo} achieves modality bridging by introducing gated cross-attention between pre-trained vision and language models; BLIP \cite{li2022blip} proposes a unified vision-language pre-training framework that supports both understanding and generation; BLIP-2 \cite{li2023blip} further employs the Q-Former to bridge a frozen visual encoder and a frozen LLM; LLaVA \cite{liu2023visual} constructs a general-purpose visual assistant through visual instruction tuning; and works such as PaLI \cite{chen2022pali}, Kosmos-1 \cite{huang2023language}, and InternVL \cite{chen2024internvl} continue to expand general multimodal capabilities and visual reasoning performance using larger-scale data and stronger components. Despite the substantial progress made by these methods in multimodal capabilities, the cost of full fine-tuning has become increasingly prohibitive as the scale of vision and language models continues to grow exponentially.

 \subsection{Parameter-Efficient Fine-Tuning for VLMs}
Parameter-Efficient Fine-Tuning techniques achieve rapid adaptation to downstream tasks by updating only a small number of learnable parameters or introducing lightweight auxiliary modules, while keeping the majority of pre-trained weights frozen. These methods have been widely adopted for task transfer in the vision \cite{rebuffi2017learning,jie2023fact,jie2023revisiting}, language model \cite{houlsby2019parameter,zaken2022bitfit,pfeiffer2021adapterfusion}, and multimodal model \cite{jiang2023mixphm,luo2023cheap,jie2024memory} domains. PEFT methods for VLMs have demonstrated competitive performance. Depending on how visual information is fused into the LLM, these methods can be broadly categorized into two paradigms. The first paradigm  (\textit{e.g.}, VL-Adapter \cite{sung2022vl}, VL-PET \cite{hu2023vl}, and LaVIN \cite{luo2023cheap}) projects image features into visual prompt tokens, concatenates them with text tokens, and feeds the resulting sequence into the LLM, while fine-tuning only the projection layer and the PEFT modules inserted into the LLM. However, these methods still suffer from efficiency bottlenecks due to redundant visual tokens. The second paradigm (\textit{e.g.}, MemVP \cite{jie2024memory} and LLaMA-Adapter v2 \cite{gao2023llama}) remove image from the input stage and instead treat it as an external visual knowledge injected into the FFN layers of the LLM. However, completely ignoring images at the input stage may lead to insufficient activation of knowledge, especially since text-only semantic modeling can introduce ambiguity in tasks that require integrating textual and visual information. Moreover, fusing visual features at the FFN layers may disrupt the activation distribution of the pre-trained language model and still struggle to address the issues of representation misalignment and activation distribution drift caused by cross-modal semantic differences.

\section{Preliminary}
\subsection{Vision-Language Model}
\paragraph{\textbf{Vision Encoder}}
The vision encoder typically employs the CLIP-ViT \cite{radford2021learning} architecture. Given an image $I\in\mathbb{R}^{H\times W\times C}$, we split it into $N=\frac{HW}{P^2}$ non-overlapping patches of size $P\times P$, where $(H,W)$ denotes the image resolution and $C$ is the number of channels. These patches are flattened into tokens and prepended with a learnable \textit{[CLS]} token. The resulting token sequence is then fed into Transformer layers to obtain visual representations.


\paragraph{\textbf{Large Language Model}}
The core of an LLM is a stack of Transformer blocks, each comprising multi-head self-attention (MHSA) and FFN, typically equipped with normalization and residual connections for stable optimization.  
Concretely, given the token $x$, a Transformer block updates it as follows: 
\begin{equation}
  \left\{
  \begin{aligned}
     x^{u} &= x + MHSA(Norm(x)), \\
     x^{u'} &= x^{u} + FFN(Norm(x^{u}))
  \end{aligned}
  \right.
\end{equation}
where $u$ is the Transformer layer number. The FFN consists of two fully connected layers and can be written as:
\begin{equation}
\mathbf{FFN}(\mathbf{x}) = \phi(\mathbf{x}\mathbf{W}_1)\mathbf{W}_2,
\label{Equa:FFN}
\end{equation}
where $W_1$ and $W_2$ are learnable projection matrices and $\phi(\cdot)$ denotes a non-linear activation.

\subsection{FFN with Visual Knowledge}
Previous work interprets FFNs in pre-trained language models as key-value memory that stores and retrieves factual associations \cite{geva2021transformer,dai2022knowledge,meng2022locating}. However, these studies have focused on purely textual models, leaving visual knowledge integration underexplored. Recently, MemVP \cite{jie2024memory} addressed this gap by encoding visual knowledge into additional keys and values within FFNs, providing language models with externally visually conditioned knowledge.

The FFN process is shown in Equation (\ref{Equa:FFN}). In fact, $W_1$ and $W_2$ can be expressed as a collection of key-value pairs:
\begin{equation}
W_1 = (k_1,k_2,\ldots,k_D), 
W_2 = (v_1,v_2,\ldots,v_D)^{\top},
\end{equation}
where $k_i \in \mathbb{R}^{d}$ and $v_i \in \mathbb{R}^{d}$ denote the $i$-th key and value, respectively. FFN performs retrieval from its key-value memory. The retrieval process for visual features could be formulated as:
\begin{equation}
\mathbf{Retrieval}(\mathbf{x})=
\sum_{i=1}^{n}\phi\!\left(\left\langle \mathbf{x},\,\mathbf{K}(\mathbf{z}_i)\right\rangle\right)\cdot \mathbf{V}(\mathbf{z}_i),
\label{Equa:FFN_re}
\end{equation}
where $K(z_i)$, $V(z_i)\in \mathbb{R}^d $ denote the key and value corresponding to the visual feature $z_i$, respectively. Accordingly, the formulation Equation (\ref{Equa:FFN_re}) shares a similar form and performs a similar process as with Equation (\ref{Equa:FFN}).

\section{Method}
We propose BTS, a novel visual enhancement framework for MLLMs, as illustrated in Figure \ref{fig:2}. \textbf{(a)} We first strengthen early visual grounding by introducing multi-scale visual context at the input stage. Specifically, \textbf{MVM} provides both coarse and fine visual information, where the coarse information capture global layout and scene structure, while the fine information preserves local details. This design allows the language model to establish visual understanding at an early stage, rather than deferring visual reasoning to deeper layers. 
\textbf{(b)} Furthermore, we improve the stability of cross-modal fusion within the FFN layers through feature calibration. Since cross-modal injection could easily perturb the activation statistics of the LLM, \textbf{AFC} performs lightweight feature calibration to keep fused activations within a well-behaved range. This makes cross-modal fusion more robust and prevents visual signals from drifting the residual stream away from the pretrained linguistic regime. 
In summary, BTS activates richer visual context early for better grounding and maintains stable fusion during FFN injection, enhancing multimodal reasoning.

\subsection{Multi-scale Visual Modulation}
\label{mvm}
Existing methods such as MemVP \cite{jie2024memory} inject visual information only within the FFN layers of the LLM, while the input stage remains purely textual. This results in a lack of explicit visual contextual priors during early processing, making it difficult to activate image-relevant factual knowledge and fine-grained visual information promptly. Consequently, later layers face greater pressure for cross-modal alignment and information retrieval. Based on this issues, we propose the Multi-scale Visual Modulation (MVM) mechanism, which extracts multi-scale visual features and converts them into a lightweight residual bias injected at the input stage, furnishing the LLM with earlier, more explicit, and richer visual priors. 

Specifically, given an image $I$, the vision encoder outputs a sequence of visual token features $Z$. We construct multi-scale visual representations by pooling adjacent tokens. Concretely, we apply pooling operations with different kernel sizes in a multi-stage downsampling operator $Down(\cdot)$ to obtain $S$ scales of visual sequences:
\begin{equation}
\{Z^{(s)}\}_{s=1}^{S}=Down(Z).
\end{equation}
The shallower scale preserves fine-grained local information, while the deeper scale emphasizes the global semantic structure. Next, for each scale, we project visual tokens into the LLM hidden space (dimension $d$) and perform
mean pooling over tokens to form a scale-wise context vector:
\begin{equation}
v^{(s)}=\frac{1}{N_s}\sum_{j=1}^{N_s}Proj\!\left(Z^{(s)}_{j}\right)\in\mathbb{R}^{B\times d}.
\end{equation}
Notably, prior methods \cite{liu2024improved,guo2024llava} typically crop or partition the original image and resize the resulting sub-images, then process each sub-image with multiple forward passes through the visual encoder to obtain multi-scale features. In contrast, our method produces multi-scale features without repeatedly invoking the visual encoder, thereby maintaining efficiency. We then aggregate multi-scale contexts to obtain a unified visual context representation $v=\frac{1}{S}\sum_{s=1}^{S}v^{(s)}$. On top of this, we introduce a \textbf{visual-bias adapter} that maps the aggregated
multi-scale visual context $v$ to a residual bias aligned with the LLM hidden space:
\begin{equation}
{\delta}=tanh(\alpha)\cdot Adapter(v),
\end{equation}
where $Adapter(\cdot)$ is a two-layer feed-forward adapter (implemented as a MLP)
that produces the visual bias to be injected into textual representations.
$\alpha$ is a learnable scalar gate, and $tanh(\cdot)$ constrains the injection strength to $(-1,1)$,
which stabilizes training and prevents overly strong visual modulation from overriding language priors. Finally, to incorporate visual information at the input stage, we add the resulting bias to text token
embeddings in a residual manner. 
Given the text token sequence $x$ with embedding matrix $x_{\mathrm{emb}}=\mathrm{Embedding}(x)\in\mathbb{R}^{B\times T\times d}$, we broadcast the multi-scale visual bias $\delta\in\mathbb{R}^{B\times 1\times d}$ along the sequence dimension and add it element-wise to obtain $x_{\mathrm{input}}=x_{\mathrm{emb}}+\delta$, where $x_{\mathrm{input}}$ denotes the input representations augmented with visual information.
A key advantage of this design is that it injects multi-scale visual contextual information at the input stage without increasing the token sequence length, thereby improving the utility of visual information for complex vision-language tasks.

\begin{table*}[t]
\caption{Accuracy on ScienceQA test set. Question categories: NAT = natural science, SOC = social science, LAN = language science, TXT = w/ text context, IMG = w/ image context, NO = no context, G1-6 = grades 1-6, G7-12 = grades 7-12.  Our results are shaded in gray.  $^{\dagger}$ Results from our reproduction of the original work. Other results are quoted from their original papers.}
\centering
\vspace{-0.2cm} 
\setlength{\tabcolsep}{8pt}
\begin{tabular}{lc|ccc|ccc|cc|c}
\toprule
\multirow{2.5}{*}{\textbf{Method}}
& \multirow{2.5}{*}{\textbf{LLM}}
& \multicolumn{3}{c|}{\textbf{Subject}} 
& \multicolumn{3}{c|}{\textbf{Context Modality}}
& \multicolumn{2}{c|}{\textbf{Grade}}
& \multirow{2.5}{*}{\textbf{Average}} \\
\cmidrule(lr){3-5}\cmidrule(lr){6-8}\cmidrule(lr){9-10}
& &\textbf{NAT} & \textbf{SOC} & \textbf{LAN}
& \textbf{TXT} & \textbf{IMG} & \textbf{NO}
& \textbf{G1-6} & \textbf{G7-12}
&  \\
\midrule
\multicolumn{10}{l}{\textit{MLLM methods}} \\
Human \cite{lu2022learn} & - & 90.23 & 84.97 & 87.48 & 89.60 & 87.50 & 88.10 & 91.59 & 82.42 & 88.40 \\
GPT-3.5 \cite{lu2022learn} & GPT-3.5 & 74.64 & 69.74 & 76.00 & 74.44 & 67.28 & 77.42 & 76.80 & 68.89 & 73.97 \\
GPT-3.5 w/ CoT \cite{lu2022learn} & GPT-3.5 &75.44 & 70.87 & 78.09 & 74.68 & 67.43 & 79.93 & 78.23 & 69.68 & 75.17 \\
GPT-4 \cite{achiam2023gpt} & GPT-4 &84.06 & 73.45 & 87.36 & 81.87 & 70.75 & 90.73 & 84.69 & 79.10 & 82.69 \\
UnifiedQA \cite{lu2022unified} & T5 &71.00 & 76.04 & 78.91 & 66.42 & 66.53 & 81.81 & 77.06 & 68.82 & 74.11 \\
MM-CoT$_{\text{Base}}$ \cite{zhang2023multimodal} & T5 &87.52 & 77.17 & 85.82 & 87.88 & 82.90 & 86.83 & 84.65 & 85.37 & 84.91 \\
LLaVA \cite{liu2024improved} & Vicuna-7B & - & - & - & - & - & - & - & - & 89.84 \\
LLaVA \cite{liu2024improved} & Vicuna-13B & 90.36 & \textbf{95.95}& 88.00 & 89.49 & 88.00 & 90.66 & 90.93 & 90.90 & 90.92 \\
\midrule
\multicolumn{10}{l}{\textit{PEFT methods}} \\
LLaMA-Adapter \cite{zhang2023llama} & LLaMA-7B &84.37 & 88.30 & 84.36 & 83.72 & 80.32 & 86.90 & 85.83 & 84.05 & 85.19 \\
LLaVA-LoRA \cite{jie2024memory} & LLaMA-7B &91.70 & 94.60 & 86.09 & 91.25 & 90.28 & 88.64 & 91.52 & 89.65 & 90.85 \\
LaVIN \cite{luo2023cheap} & LLaMA-7B &89.25 & 94.94 & 85.24 & 88.51 & 87.46 & 88.08 & 90.16 & 88.07 & 89.41 \\
MemVP \cite{jie2024memory} & LLaMA-7B &94.45 & 95.05 & 88.64 & 93.99 & 92.36 & 90.94 & 93.10 & 93.01 & 93.07 \\
ADEM-VL$^{\dagger}$ \cite{hao2025adem} & LLaMA-7B &94.49 & 95.28 & 87.64 & 94.09 & 93.06 & 90.10 & 93.10 & 92.49 & 92.88 \\
IPK \cite{song2026leveraging} & LLaMA-7B & - & - & - & - & - & - & - & - & 92.30 \\
\rowcolor{gray!15}
Ours & LLaMA-7B &94.80 & 95.28 & 89.73 & 94.48 & 93.11 & 91.78 & 93.72 & 93.34 & 93.59 \\
LaVIN \cite{luo2023cheap} & LLaMA-13B & 90.32 & 94.38 & 87.73 & 89.44 & 87.65 & 90.31 & 91.19 & 89.26 & 90.50 \\
MemVP \cite{jie2024memory} & LLaMA-13B & 95.07 & 95.15 & 90.00 & 94.43 & 92.86 & \textbf{92.47} & 93.61 & 94.07 & 93.70 \\
\rowcolor{gray!15}
Ours & LLaMA-13B & \textbf{95.65} & 95.83 & \textbf{92.82} & \textbf{95.67} & \textbf{93.21} & 92.29 & \textbf{94.17} & \textbf{94.10} & \textbf{94.35} \\
\bottomrule
\end{tabular}
\label{tab:1}
\vspace{-0.2cm}
\end{table*}

\subsection{Adaptive Fused-feature Calibration}
\label{afc}
Based on the perspective that FFN can be viewed as key-value memory stores for factual knowledge \cite{geva2021transformer}, we treat visual knowledge as new memory entries and inject it into FFN layers to compensate for the lack of vision-related knowledge in language models. Concretely, from the FFN standpoint, visual features are converted into additional key-value pairs $(K(z_i), V(z_i))$ and written into the FFN retrieval space. This augments the original language-fact memory with visually conditioned knowledge support, expanding the capacity of the model for vision-grounded facts. After inserting visual memory entries, the output of the FFN can be formulated as the joint retrieval of both linguistic and visual memories:
\begin{equation}
FFN(x)
=\sum_{i=1}^{D}\phi\!\left(\langle x, k_i\rangle\right)\cdot v_i
+\sum_{i=1}^{n}\phi\!\left(\langle x, K(z_i)\rangle\right)\cdot V(z_i),
\end{equation}
where $z_i$ denotes visual feature, and $K(z_i)$ and $V(z_i)$ are its corresponding visual key and value, respectively. 
However, naively injecting visual knowledge into the FFN layers may lead to representation misalignment and activation distribution drift due to cross-modal semantic discrepancies. To this end, we propose an Adaptive Fused-feature Calibration (AFC) module, which performs per-channel affine transformation on the fused activations to adaptively calibrate their distribution. This decouples visual injection from distribution alignment, thereby improving the stability and controllability of the fusion process. 

Beyond the standard FFN formulation commonly used in small to medium-scale language models, larger models often adopt Gated Linear Units (GLU) \cite{touvron2023llama} to enhance the FFN’s expressive capacity. In practice, we adopt a two-sided SiLU gating mechanism to facilitate feature interaction in the FFN, formulated as follows:
\begin{equation}
\begin{aligned}
FFN(x)
=
&\sum_{i=1}^{D}
\big\langle SiLU(x),\, SiLU(k_i) \big\rangle
\cdot v_i \\
&+\sum_{i=1}^{n}
\big\langle SiLU(x),\, SiLU(K(z_i)) \big\rangle
\cdot V(z_i).
\end{aligned}
\end{equation}

To further mitigate the feature distribution shift caused by cross-modal injection, AFC applies a \textbf{channel-wise affine calibration} to the fused activations as follows:
\begin{equation}
AFC(x)={\gamma}\odot x+{\beta},
\end{equation}
where $x$ denotes the joint retrieval of both linguistic and visual memories, ${\gamma}$ and ${\beta}$ are learnable per-channel scale and bias parameters, respectively, and $\odot$ denotes element-wise multiplication. In general, AFC calibrates post-fusion activations to regulate their statistics, providing a lightweight yet effective mechanism for stabilizing the semantic stream of pretrained LLMs.

In our implementation, we apply AFC twice along the vision-conditioned FFN pathway, aligned with the two projections of the FFN (down- and up-projection). We calibrate after the down-projection because it compresses information into a bottleneck representation, whose activation statistics are particularly susceptible to cross-modal semantic discrepancies. Early calibration stabilizes intermediate magnitudes and suppresses distribution drift before it is amplified by subsequent nonlinear transformations and residual accumulation. We then apply the second AFC at the FFN output to re-align the fused activations to a numerically stable regime compatible with the pretrained LLM residual stream. Hence, the vision-conditioned FFN with two-stage AFC is computed as:

\begin{equation}
  \left\{
  \begin{aligned}
     y' 
&= AFC\left(SiLU(x)\cdot SiLU(W_{1})^{\top}\right), \\
y 
&= AFC\left(y'\cdot W_{2}\right),
  \end{aligned}
  \right.
\end{equation}

where $W_{1}$ and $W_{2}$ denote the down- and up-projection matrices in the FFN, respectively, and $y'$ represents the intermediate activation retrieved by the FFN. For AFC, we initialize ${\gamma}$ and ${\beta}$ with small perturbations so that AFC begins close to an identity mapping. This minimizes disruption to pretrained language behavior during early optimization, while enabling AFC to progressively learn modality-aware scaling and shifting for robust distribution alignment.

\begin{table}[t]
\caption{Comparative performance evaluation on multiple VQA benchmarks. Best result is in bold face.}
\vspace{-0.3cm}
\centering
\setlength{\tabcolsep}{3pt}
\begin{tabular}{l|ccccc}
\toprule
\textbf{Method} & \textbf{GQA} & \textbf{VQAv2} & \textbf{OKVQA} & \textbf{VizWiz}  & \textbf{SQA} \\
\midrule
\multicolumn{6}{l}{\textit{MLLM methods}} \\
BLIP-2 \cite{li2023blip}          & 41.0 & 41.0 & 45.9 & 19.6 & 74.2 \\
InstructBLIP-8B \cite{dai2023instructblip}  & 49.2 & 51.9 & 49.9 & 34.5 & 73.3 \\ 
IDEFICS \cite{laurenccon2023obelics}          & 38.4 & 50.9 & 38.4 & 35.5 & 77.4 \\
LLaVA \cite{liu2023visual}           & 49.5 & 63.6 & 45.2 & 30.7 & 56.5 \\
LLaVA-v1.5 \cite{liu2024improved}      & 62.0 & 74.0 & 45.6 & 50.0 & 90.9 \\
InternVL-MLP \cite{chen2024internvl}    & 62.9 & 79.3 & 42.9 & 52.5 & 90.1 \\
InternVL-QLLaMA \cite{chen2024internvl} & 57.7 & 72.3 & 51.0 & 44.5 & 88.6 \\
Qwen2-VL \cite{wang2024qwen2}        & 63.7 & \textbf{79.8} & 48.5 & 65.4 & 79.1 \\
\midrule
\multicolumn{6}{l}{\textit{PEFT methods}} \\
LLaMA \cite{touvron2023llama}           & 34.0 & 40.1 & 27.1 & 50.3 & 36.2  \\ 
LLaMA-Adapter v2 \cite{gao2023llama}  & 33.8 & 41.6 & 37.8 & 42.2 & 85.2 \\
LaVIN \cite{luo2023cheap}     & 55.2 & 68.6 & 46.1 & 59.6 & 89.4 \\
MemVP \cite{jie2024memory}           & 58.1 & 73.2 & 49.9 & 70.0 & 93.1 \\
IPK \cite{song2026leveraging} & 66.2 & 75.6 & 51.1 & 71.4 & 92.3 \\
\rowcolor{gray!15}
Ours             & \textbf{66.3} & 77.6 & \textbf{52.5} & \textbf{72.5} & \textbf{93.6} \\
\bottomrule
\end{tabular}
\label{tab:2}
\vspace{-0.2cm}
\end{table}

\section{Experiments}
\subsection{Dataset and Experiment Setup}
We conduct experiments primarily on ScienceQA \cite{lu2022learn}, which contains 21k multimodal multiple choice questions with rich domain diversity across 3 subjects, 26 topics, 127 categories, and 379 skills. The dataset is split into training,
validation, and test splits with 12726, 4241, and 4241 examples, respectively, and we report average accuracy on the test set. In addition, following prior work \cite{luo2024feast,liu2024improved}, we further evaluate our method on multiple VQA benchmarks, including GQA \cite{hudson2019gqa}, VQAv2 \cite{goyal2017making}, OKVQA \cite{marino2019ok}, and VizWiz \cite{gurari2018vizwiz}. We use LLaMA-7B and LLaMA-13B \cite{touvron2023llama} as the language backbones, together with CLIP ViT-L/14 \cite{radford2021learning} as the vision encoder. Following the training setup of MemVP \cite{jie2024memory}, we train the models for 20 epochs using the AdamW optimizer, with a batch size of 32 and a learning rate of 9e-3. The positional embedding length is set to 320 for LLaMA-7B and 400 for LLaMA-13B. For samples without image inputs, we set the visual prompts to zero tensors. The learnable scalar gate \(\alpha\) is initialized to 0, and \(\tanh(\cdot)\) limits the injection strength to \((-1, 1)\), which stabilizes training and prevents visual modulation from overriding language priors. All experiments are conducted in a single-stage end-to-end setting, where answers are generated directly from the inputs. Experiments are implemented in PyTorch 2.0 and conducted on RTX A800 GPU. 


\subsection{Comparison Methods}
We compare with two categories of methods: (1) MLLM methods. We compare our method with representative vision-language models spanning diverse architectures, including BLIP-2 \cite{li2023blip}, Flamingo \cite{alayrac2022flamingo}, InstructBLIP \cite{dai2023instructblip}, IDEFICS \cite{laurenccon2023obelics}, UnifiedQA \cite{lu2022unified}, LLaVA/LLaVA-1.5 \cite{liu2023visual,liu2024improved}, InternVL \cite{chen2024internvl}, Qwen-VL/Qwen2-VL \cite{wang2024qwen2}, GPT-3.5 \cite{lu2022learn}, GPT-4 \cite{achiam2023gpt} and MM-CoT \cite{zhang2023multimodal}. (2) PEFT methods. We further compare our method with parameter-efficient fine-tuning methods, including LLaMA-Adapter \cite{gao2023llama}, LLaVA-LoRA \cite{jie2024memory}, LaVIN \cite{luo2023cheap}, MemVP \cite{jie2024memory}, ADEM-VL \cite{hao2025adem}, and IPK \cite{song2026leveraging}. 



\subsection{Results Analysis}
\paragraph{\textbf{Results on ScienceQA}} As shown in Table~\ref{tab:1}, under the same LLaMA-7B backbone, our method achieves state-of-the-art performance on ScienceQA, reaching an average accuracy of 93.59\%. It outperforms strong PEFT baselines, including MemVP (93.07\%), ADEM-VL (92.88\%), LaVIN (89.41\%), LLaVA-LoRA (90.85\%), and LLaMA-Adapter (85.19\%). When scaled to LLaMA-13B, our method further improves the average accuracy to 94.35\%, again surpassing the existing baselines. Particularly, under the 7B setting, our method improves over MemVP by 0.35\%, 0.23\%, and 1.09\% on NAT, SOC, and LAN, respectively, and also achieves gains of 0.49\% and 0.75\% on the TXT and IMG evaluations. Similar trends are observed under the 13B setting, where the improvements are particularly pronounced on LAN and TXT. This suggests that our method not only provides richer and earlier visual contextual support, but also effectively mitigates distributional perturbations introduced by cross-modal fusion, improving overall multimodal understanding while preserving language capabilities.  

\paragraph{\textbf{Results on Multiple VQA Benchmarks}} Furthermore, to evaluate the generalization capability of the proposed method, we conduct experiments on multiple VQA benchmarks, as shown in Table~\ref{tab:2}. Our method achieves competitive performance on GQA, OKVQA, and VizWiz, achieving 66.3\%, 52.5\%, and 72.5\%, respectively. In particular, the improvements on GQA and VizWiz indicate that our method is effective not only in modeling visual-semantic relationships in complex scenes, but also in handling low-quality and noisy real-world images more robustly. Overall, these results further demonstrate that our method generalizes well on ScienceQA and across diverse VQA tasks.

\begin{table}[t]
\caption{Ablation study on each component. $down$ and $up$ represent two projections of FFN, respectively.}
\vspace{-0.3cm}
\centering
\setlength{\tabcolsep}{8pt}
\begin{tabular}{c|c|cc|c|c|c}
\toprule
\multirow{2}{*}{\#} & \multirow{2}{*}{MVM} & \multicolumn{2}{c|}{AFC} &
\multirow{2}{*}{OKVQA} & \multirow{2}{*}{VizWiz} & \multirow{2}{*}{SQA} \\
\cline{3-4}
 &  & $down$ & $up$ &  &  &  \\
\midrule
0 & $\checkmark$ & - & - & 52.1 & 70.8 & 93.2 \\
1 & - & $\checkmark$ & $\checkmark$ & 51.1 & 70.1 & 92.9 \\
2 & $\checkmark$ & $\checkmark$ & - & 51.5 & 71.9 & 93.3 \\
3 & $\checkmark$ & - & $\checkmark$ & 51.3 & 71.3 & 92.8 \\
4 & $\checkmark$ & $\checkmark$ & $\checkmark$ & 52.5 & 72.5 & 93.6 \\ 
\bottomrule
\end{tabular}
\label{tab:3}
\vspace{-0.2cm}
\end{table}

\begin{figure*}[t]
  \centering
  \includegraphics[width=\linewidth,trim=25 370 220 15,clip]{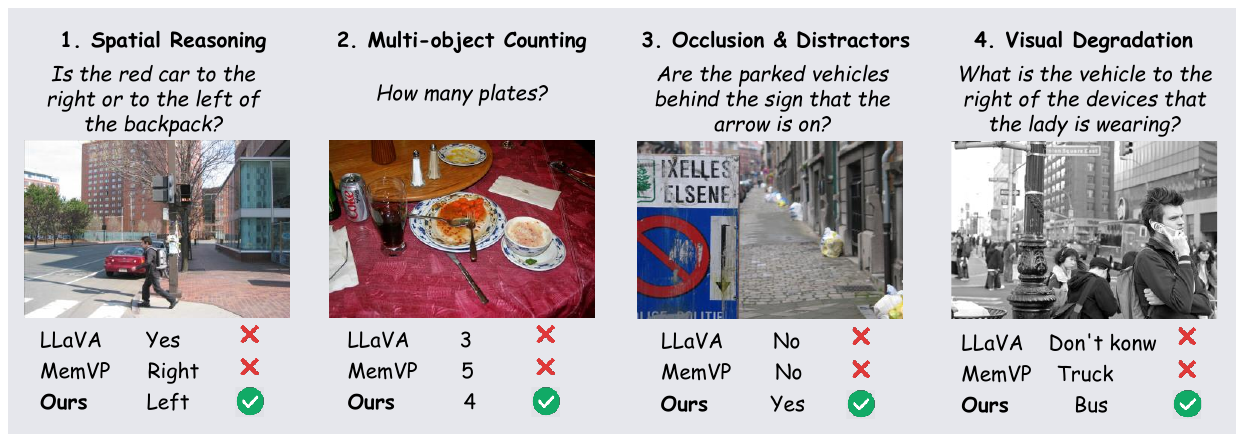} 
  \vspace{-0.7cm}
  \caption{Case study comparing our method with other methods on challenging visual reasoning tasks.}
  \label{fig:case}
  \vspace{-0.2cm}
\end{figure*}

\begin{table}[t]
\caption{Ablation study on  Multi-scale Visual Modulation (MVM) scale settings.}
\vspace{-0.3cm}
\centering
\setlength{\tabcolsep}{14pt}
\begin{tabular}{l|c|c|c}
\toprule
MVM & OKVQA & VizWiz & SQA \\
\midrule
$[256]$ & 51.5 & 71.8 & 93.2 \\
$[256,64]$ & 52.5 & 72.5 & 93.6 \\
$[256,64,16]$ & 51.8 & 72.0 & 93.1 \\
\bottomrule
\end{tabular}
\label{tab:4}
\vspace{-0.2cm}
\end{table}

\begin{table}[t]
\centering
\caption{Ablation study on different LLM backbones on SQA.}
\vspace{-0.3cm}
\renewcommand{\arraystretch}{1.12}
\setlength{\tabcolsep}{11pt}

\begin{tabular}{lccc}
\toprule
\multicolumn{1}{c|}{\multirow{2}{*}{\textbf{Method}}} 
& \multicolumn{2}{c|}{\textbf{LLM backbone}} 
& \multirow{2}{*}{\textbf{SQA}} \\
\cline{2-3}
\multicolumn{1}{c|}{} 
& \multicolumn{1}{c|}{\textbf{LLaMA-7B}} 
& \multicolumn{1}{c|}{\textbf{Vicuna-7B}} 
& \\
\midrule
\multicolumn{1}{l|}{ADEM-VL}  
& \multicolumn{1}{c|}{$\checkmark$} 
& \multicolumn{1}{c|}{-} 
& 92.88 \\

\multicolumn{1}{l|}{MemVP} 
& \multicolumn{1}{c|}{$\checkmark$} 
& \multicolumn{1}{c|}{-} 
& 93.07 \\

\multicolumn{1}{l|}{Ours}  
& \multicolumn{1}{c|}{$\checkmark$} 
& \multicolumn{1}{c|}{-} 
& 93.59 \\

\multicolumn{1}{l|}{ADEM-VL}  
& \multicolumn{1}{c|}{-} 
& \multicolumn{1}{c|}{$\checkmark$} 
& 91.75 \\

\multicolumn{1}{l|}{MemVP} 
& \multicolumn{1}{c|}{-} 
& \multicolumn{1}{c|}{$\checkmark$} 
& 91.26 \\

\multicolumn{1}{l|}{Ours}  
& \multicolumn{1}{c|}{-} 
& \multicolumn{1}{c|}{$\checkmark$} 
& 92.51 \\
\bottomrule
\end{tabular}
\label{tab:5}
\vspace{-0.2cm}
\end{table}

\subsection{Ablation Studies}
\paragraph{\textbf{Ablation on Each Component}} To validate the effectiveness of each component, we conduct ablation studies on OKVQA, VizWiz, and SQA. As shown in Table \ref{tab:3}, introducing only MVM (\# 0) achieves 52.1\%, 70.8\%, and 93.2\% on OKVQA, VizWiz, and SQA, respectively, indicating that introducing multi-scale visual context at the input stage helps activate image-related knowledge. In contrast, using only AFC (\# 1) provides relatively limited gains. Moreover, based on MVM, employing only unilateral calibration (\# 2 or \# 3) is inferior to bilateral calibration, suggesting that AFC plays complementary stabilizing roles at two critical stages of FFN-based visual retrieval. As a result, combining MVM with AFC (\# 4) achieves the best results, further validating the effectiveness of the proposed approach. 

\paragraph{\textbf{Ablation on MVM}} We further ablate the scale settings of MVM. As shown in Table~\ref{tab:4}, using only fine-grained features (\textit{i.e.}, [256]) performs worse than combining fine-grained and coarser context (\textit{i.e.}, [256, 64]), indicating that the multi-scale priors introduced by MVM provide richer visual context and help the model leverage both coarse-grained semantics and fine-grained information. Adding an additional coarser scale (\textit{i.e.}, [256, 64, 16]) slightly degrades performance, suggesting that introducing too many scales may add redundancy.

\paragraph{\textbf{Ablation on LLM Backbone}} To evaluate the adaptability of our method across different LLM backbones, we conduct comparative experiments using LLaMA-7B and Vicuna-7B. As reported in Table~\ref{tab:5}, our method achieves the best performance under both settings, attaining 93.59\% on LLaMA-7B and 92.51\% on Vicuna-7B, respectively. Compared with ADEM-VL and MemVP, our method improves performance on both backbones, demonstrating its robustness across diverse LLM backbones. It is also worth noting that the performance on Vicuna-7B is slightly lower than that on LLaMA-7B. A possible explanation is that Vicuna, which undergoes additional instruction fine-tuning based on LLaMA, is optimized for dialogue alignment and instruction following, and may therefore not fully preserve the original representation space that proves more suitable for visual question answering tasks.


\subsection{Case Study}
Figure~\ref{fig:case} provides four qualitative examples. (1) Spatial reasoning: when judging whether the red car is to the left or right of the backpack, LLaVA and MemVP are misled by contextual distractors, while our method correctly infers the relative position by establishing the global layout early. (2) Multi-object counting: in cluttered scenes, prior methods miss or double-count objects, whereas our method produces the correct count. (3) Occlusion and distractors: under heavy occlusion, our method localizes the key evidence and makes the correct prediction. (4) Visual degradation: in low-quality blurred grayscale images, our method still captures appropriate visual cues to correctly identify the vehicle as a “Bus.” In general, these cases suggest that our method can more reliably retrieve and utilize visual information under challenging conditions.

\begin{figure}[t]  
  \centering
  \includegraphics[width=\linewidth,trim=25 270 370 30,clip]{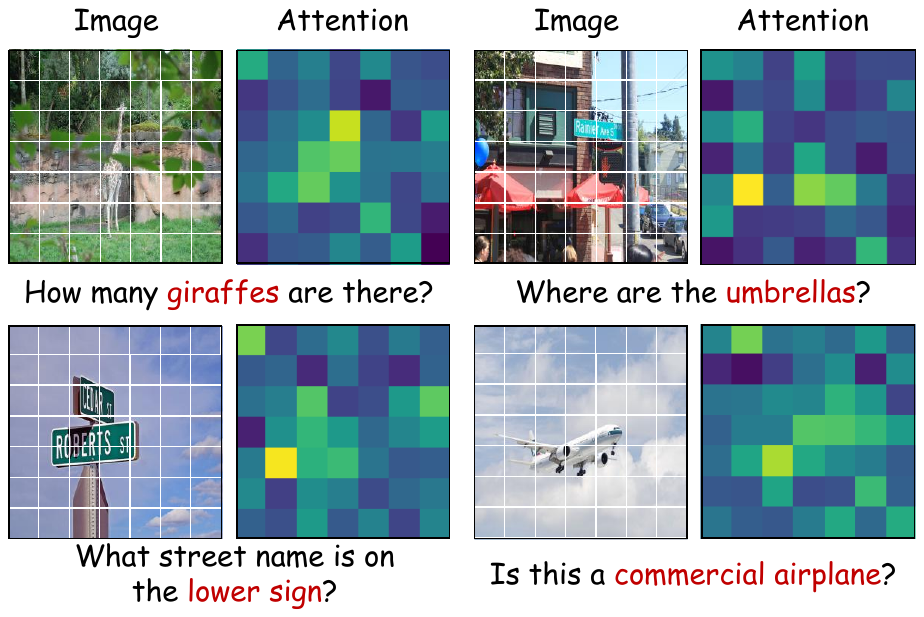} 
  \vspace{-0.7cm}
  \caption{Visualization of attention on image patches.} 
  \label{fig:attn}
  \vspace{-0.2cm}
\end{figure}

\begin{figure*}[t] 
\centering
\includegraphics[width=\linewidth,trim=40 263 720 50,clip]{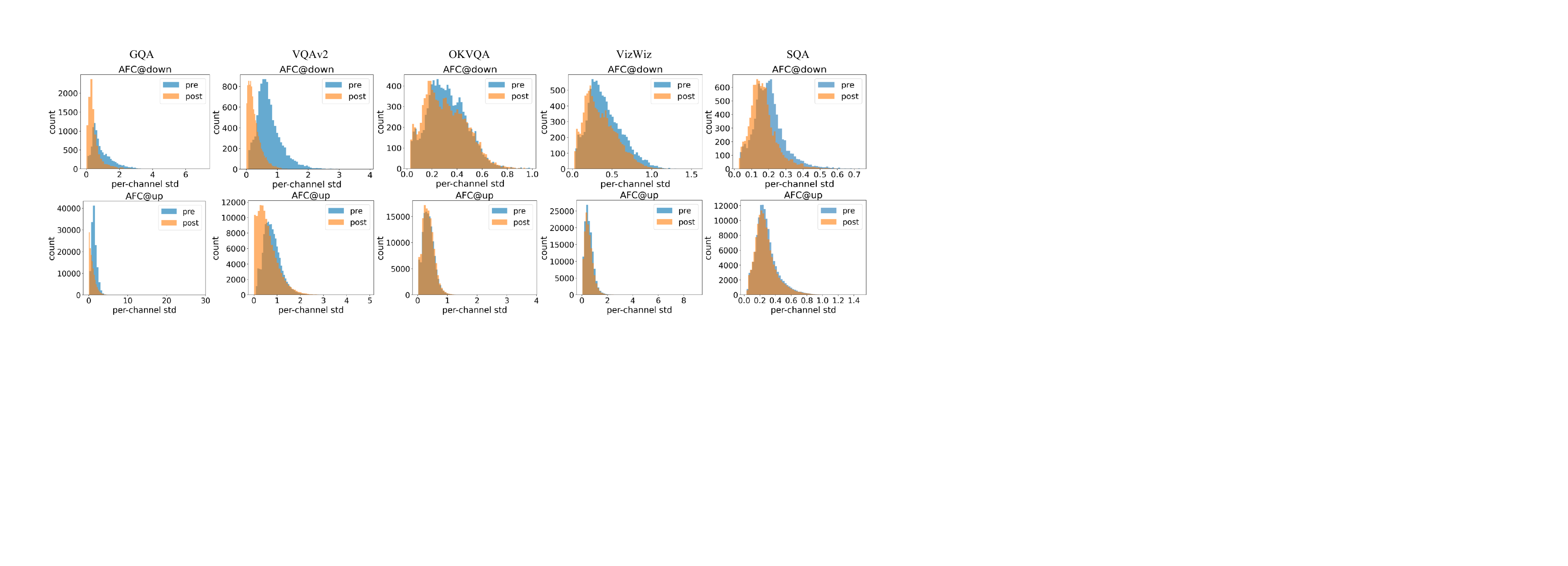}   
\color{blue}
\vspace{-0.6cm}
\caption{Visualization of the distributions of per-channel activation standard deviation before (pre) and after (post) AFC.}
\label{fig:afc}
\vspace{-0.2cm}
\end{figure*}

\subsection{Attention Visualization}
Figure \ref{fig:attn} visualizes the attention distribution on visual tokens in the final layer of the LLM when addressing VQA multimodal questions. As shown in the visualization, for counting queries (\textit{e.g.}, “How many giraffes are there?”), the highest attention concentrates on the local grid cells containing the giraffes, exhibiting a clear instance-centric focusing behavior. In contrast, for localization queries (\textit{e.g.}, “Where are the umbrellas?”), the attention forms a contiguous, band-like pattern aligned with the umbrellas’ contours, clearly delineating the target’s spatial extent even under cluttered backgrounds. Similarly, for the question “what street name is on the lower sign?”, the attention is highly concentrated on the lower sign region. For the question “Is this a commercial airplane?”, the attention mainly covers the airplane itself rather than the background. These observations collectively suggest that our method can improve vision-language alignment across diverse visual reasoning scenarios.


\subsection{Analysis of AFC}
We visualize the distributions of per-channel activation standard deviation before and after AFC at the two FFN projections (down and up) across five VQA benchmarks (\textit{i.e.}, GQA, VQAv2, OKVQA, VizWiz, and SQA), as shown in Figure \ref{fig:afc}. The blue bars (“pre”) represent the distributions before applying AFC, while the orange bars (“post”) represent those after AFC. It can be observed that the calibration effect is most pronounced at AFC@down. Across all datasets, the post-AFC distribution shifts toward lower values and becomes more concentrated than the pre-AFC distribution. This shift indicates that the original distributions exhibit heavy tails with high‑variance channels. AFC effectively suppresses these tails, demonstrating its ability to dampen abnormal responses induced by cross‑modal injection and to reduce channel‑wise variance imbalance, thereby limiting the propagation of distribution drift. In contrast, AFC@up shows substantial overlap between the pre- and post-AFC curves, with only minor adjustments. This suggests that the output‑side calibration is conservative, primarily aimed at bringing the fused features back into a numerically stable range compatible with the pretrained residual stream, preventing small mismatches from accumulating across layers and eroding linguistic priors. Overall, these results indicate that AFC decouples visual injection from statistical alignment and stabilizes cross-modal feature fusion across diverse benchmarks.


\begin{figure}[h] 
  \centering
  \includegraphics[width=\linewidth,trim=0 0 0 5,clip]{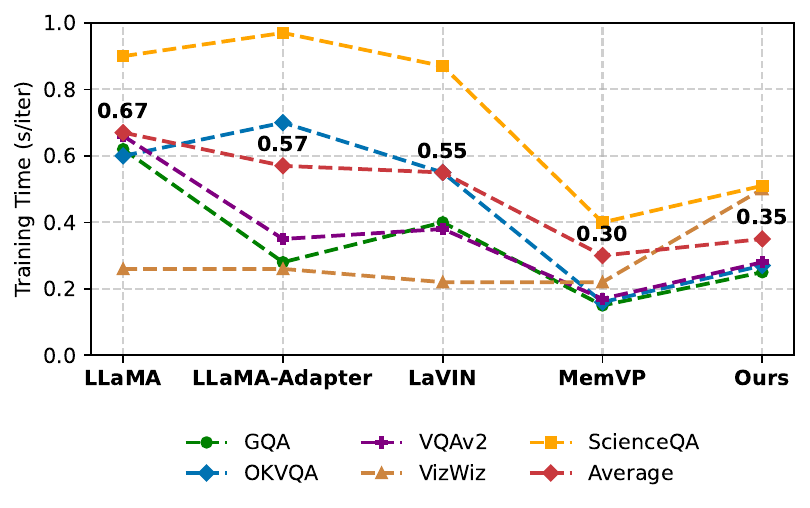} 
  \vspace{-0.7cm}
  \caption{Average training time (s/iter) across tasks.} 
  \label{fig:time}
  \vspace{-0.2cm}
\end{figure}

\subsection{Training Time  Comparison}
\label{Training Time  Comparison}
Figure \ref{fig:time} illustrates the training time of various methods across different tasks, while Table \ref{tab:2} presents the corresponding accuracy results. Our method is substantially more efficient than LLaMA, LLaMA-Adapter, and LaVIN, while achieving training time comparable to MemVP. This efficiency stems from avoiding redundant visual tokens, introducing early visual priors through lightweight modulation, and stabilizing feature fusion through calibration, thereby improving visual reasoning without noticeable overhead and making the approach more practical for deployment.

\section{Conclusion} 
In this paper, we propose Beyond Token Stacking (BTS), a visual enhancement framework that tackles two limitations in current MLLMs: (i) redundant visual tokens and high computation from token concatenation, and (ii) FFN-based visual knowledge injection that lacks early visual context and could cause activation distribution drift. BTS comprises two key components. First, Multi-scale Visual Modulation (MVM) extracts multi-scale visual context from vision encoder outputs and injects it as a lightweight residual bias into text embeddings, enabling earlier visual grounding without introducing extra visual tokens. Second, Adaptive Fused-feature Calibration (AFC) calibrates fused activations via per-channel affine transformations, stabilizing FFN-based visual injection and improving fusion robustness. Extensive experiments across multiple multimodal benchmarks demonstrate that BTS substantially enhances both multimodal understanding and reasoning capabilities.

\begin{acks}
This work was supported by the National Key Research and Development Program under Grant 2024YFB4506200, the Science and Technology Innovation Program of Hunan Province under Grant 2024RC1048, the Key Research and Development Program of Hunan Province under Grant 2026QK3002.
\end{acks}
\vspace{-0.2cm}

\input{sample-sigconf-authordraft.bbl}

\end{document}